\documentclass[a4paper,twoside]{article}

\usepackage{epsfig}
\usepackage{subcaption}
\usepackage{calc}
\usepackage{amssymb}
\usepackage{amstext}
\usepackage{amsmath}
\usepackage{amsthm}
\usepackage{multicol}
\usepackage{pslatex}
\usepackage{apalike}
\usepackage{algorithm2e}
\usepackage[bottom]{footmisc}
\usepackage{SCITEPRESS}

\begin{document}

\title{Financial time series augmentation using \\ transformer based GAN architecture}

\author{\authorname{Andrzej Podobiński\sup{1}\orcidAuthor{0009-0003-7574-2922}, Jarosław A. Chudziak\sup{1}\orcidAuthor{0000-0003-4534-8652}}
\affiliation{\sup{1}Faculty of Electronics and Information Technology, Warsaw University of Technology, Warsaw, Poland}
\email{andrzej.podobinski.stud@pw.edu.pl, jaroslaw.chudziak@pw.edu.pl}
}


\abstract{Time-series forecasting is a critical task across many domains, from engineering to economics, where accurate predictions drive strategic decisions. However, applying advanced deep learning models in challenging, volatile domains like finance is difficult due to the inherent limitation and dynamic nature of financial time series data. This scarcity often results in sub-optimal model training and poor generalization. The fundamental challenge lies in determining how to reliably augment scarce financial time series data to enhance the predictive accuracy of deep learning forecasting models. Our main contribution is a demonstration of how Generative Adversarial Networks (GANs) can effectively serve as a data augmentation tool to overcome data scarcity in the financial domain. Specifically, we show that training a Long Short-Term Memory (LSTM) forecasting model on a dataset augmented with synthetic data generated by a transformer-based GAN (TTS-GAN) significantly improves the forecasting accuracy compared to using real data alone. We confirm these results across different financial time series (Bitcoin and S\&P500 price data) and various forecasting horizons. Furthermore, we propose a novel, time series specific quality metric that combines Dynamic Time Warping (DTW) and a modified Deep Dataset Dissimilarity Measure (DeD-iMs) to reliably monitor the training progress and evaluate the quality of the generated data. These findings provide compelling evidence for the benefits of GAN-based data augmentation in enhancing financial predictive capabilities.}

\onecolumn \maketitle \normalsize \setcounter{footnote}{0} \vfill
\section{\uppercase{Introduction}}
\label{sec:introduction}
Time-series forecasting is a foundational and critical task that helps in high-stakes operational decisions across diverse domains, from optimizing energy grids to predicting financial market volatility \cite{hyndman2018forecasting}. In recent years, deep learning models have emerged as the state-of-the-art method for these complex challenges, often outperforming traditional statistical approaches and setting new benchmarks for the forecasting task \cite{lu2025survey}. This progress is especially important within financial time-series analysis, where even marginal improvements in forecasting accuracy translate directly into significant profitability gains.

\begin{figure}
    \centering
    \includegraphics[width=1\linewidth]{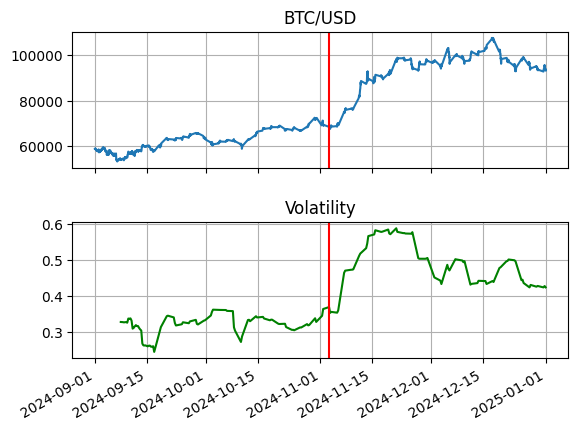}
    \caption{Bitcoin price and volatility before and after US president election 2024}
    \label{fig:btc_volatility}
\end{figure}

Despite their performance, deep learning models are inherently data-hungry. Their performance is highly dependent on dataset size, and training them with insufficient data severely limits their utility \cite{bhatt2024data}. This issue is acutely pronounced in domains characterized by scarcity and high dynamism, such as financial markets. Financial time series are inherently volatile and characterized by evolving, non-stationary market regimes \cite{mills2008econometric}, as demonstrated in Fig. \ref{fig:btc_volatility}. This results in a short time horizons of data relevant for model training.  When faced with such limited data, complex deep learning architectures become prone to overfitting \cite{kaplan2020scaling}. 

The key question this research aims to address is: Can a transformer based GAN be successfully used for data augmentation in order to improve the generalization and predictive capabilities of deep learning models when applied to volatile financial time series? We hypothesize that a data augmentation using TTS-GAN architecture, can effectively mitigate the limitations of data scarcity in financial time series by generating synthetic samples that capture the underlying statistical and temporal properties of the real data.

To verify this hypothesis, we present comprehensive experiment that compares performance of a standard Long Short-Term Memory (LSTM) \cite{hochreiter1997long} forecasting model trained only on real data against the same model trained on an augmented dataset. We demonstrate that augmentation leads to a reduction in the Mean Squared Error (MSE), validating the GAN approach for financial time series. 

Our contributions include the demonstration of a TTS-GAN as an effective data augmentation tool for overcoming scarcity in financial time series. We introduce technical enhancements to the TTS-GAN architecture, specifically utilizing optimized PyTorch modules and a "Simplified Gradient Penalty"\cite{mescheder2018training} to ensure stable convergence on volatile data. Additionally, we propose a quality metric, DTW DeD-iMs, which combines Dynamic Time Warping and Deep Dataset Dissimilarity Measure to reliably evaluate both the temporal fidelity and distributional similarity of synthetic sequences.

The remainder of this paper is organized as follows: Section 2 reviews related work regarding generative models for time series and data augmentation in the financial domain. Section 3 details our approach, including technical enhancements to the TTS-GAN architecture and the introduction of the DTW DeD-iMs convergence metric. Section 4 describes the experimental design and the datasets preparation. Section 5 presents and discusses the results of the forecasting improvements , and Section 6 concludes the paper with a summary of findings and future research directions.

\section{\uppercase{Related works}}

In the domain of deep learning for time series analysis, the scarcity of sufficiently large and representative datasets poses a significant challenge, as highlighted by Wen et al. \cite{wen2020time}. Effective augmentation methods must not only generate diverse samples but also accurately mimic the underlying temporal dependencies and statistical characteristics of real-world data. In this section, we review different GAN architectures for time series found in the recent literature and the state of augmentation in financial data. Additionally, we discuss challenges in GAN training and detail architectural and algorithmic enhancements for stable GAN convergence.

\subsection{Generative Models for Time Series}
While GANs have garnered considerable attention, the challenge of generating high-fidelity time series data is distinct from image or text generation. Early works, such as Recurrent GAN (RCGAN) and TimeGAN \cite{yoon2019time} , leveraged Recurrent Neural Networks (RNNs) to capture temporal dynamics. RNN-based approaches, however, often struggle with long-range dependencies and maintaining coherence across extended time sequences due to the inherent limitations of standard RNNs \cite{esteban2017real} \cite{pascanu2013difficulty}. While TimeGAN \cite{yoon2019time} partially mitigates this by incorporating a supervised loss component, these architectures are fundamentally constrained by the sequential processing of RNNs, limiting their ability to effectively model the complex, multi-scale dependencies characteristic of non-stationary financial data. 

The emergence of the Transformer architecture has provided an efficient mechanism for modeling these long-range dependencies through the self-attention mechanism. The TTS-CGAN (Transformer Time-Series Conditional GAN) \cite{li2022ctts}, based on TTS-GAN \cite{li2022tts} initially demonstrated on biosignal data, adopts this strength by building its generator and discriminator components entirely on transformers. This architectural choice is a fundamental advantage over RNN-based GANs, as it allows the model to process the entire sequence in parallel and more effectively capture the intricate, non-linear relationships and temporal dependencies in  financial timeseries datasets.

 \begin{figure}
     \centering
     \includegraphics[width=1\linewidth]{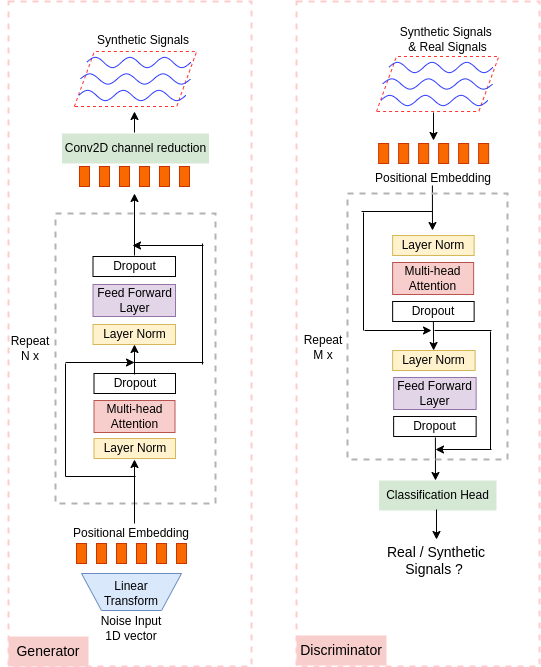}
     \caption{TTS-GAN architecture\cite{li2022tts}}
     \label{fig:placeholder}
 \end{figure}

\subsection{Augmentation for financial datasets}
Research confirming the utility of data augmentation in the financial domain provides a strong foundation for our work. A key study by Fons et al. \cite{fons2020evaluating} alongside the comprehensive survey by Iwana and Uchida \cite{iwana2021empirical} emphasizes that augmentation is a powerful, model-independent strategy. The authors specifically address the issue of limited training data, noting that augmentation is highly effective for reducing overfitting and enhancing the generalization capabilities of neural networks. To establish their baseline, Fons et al. used an LSTM model and the S\&P500 dataset, which makes their experimental context highly relevant to our research.

Crucially, while these prior works validate the overall benefit of augmentation in time series and compares various methods, they do not include a key analysis relevant to our approach. Specifically, neither the research by Fons et al. \cite{fons2020evaluating} nor the research by Iwana and Uchida \cite{iwana2021empirical} provides results for augmentation performed using the GAN framework. Our study aims to address this gap by exploring the impact of GAN-based augmentation on stock forecasting accuracy.

\subsection{Model convergence and algorithmic enhancement }
One of the challenges pronounced in the literature is the lack of consensus on metrics for evaluating generated data. This issue is particularly visible in the time-series domain, which remains comparatively less explored \cite{brophy2023generative}. Jeng et al. \cite{jeng2025generative} recently recognized similar challenges when applying GANs for the augmentation of time-series data in electric vehicle battery state-of-charge modeling. Traditional measures, such as the Wasserstein distance, assess the overall data distribution but fail to adequately penalize dissimilarities of temporal ordering or phase alignment \cite{cohen2021aligning}. Our work directly addresses this gap by proposing a novel evaluation method that combines DTW \cite{berndt1994using} and DeD-iMs \cite{calderon2022dataset}. This combination creates a measurement method that can account for temporal similarity and overall data distribution. The choice of DTW for quantifying temporal similarity was based on multiple successful applications of this method over recent decades \cite{sakoe2003dynamic}\cite{bartolini2005warp}\cite{kholmatov2005identity}\cite{prabhu2024generative}\cite{jeng2025generative}.

Furthermore, our contribution extends to resolving an architectural limitation of the original TTS-GAN architecture \cite{li2022tts}. We improve the model by integrating a performance-optimized PyTorch-based Transformer block \cite{paszke2019pytorch}. This change is coupled with the utilization of an enhanced training algorithm designed to achieve better convergence stability \cite{mescheder2018training}, thereby contributing to more reliable and high-quality synthetic time-series generation.

\section{\uppercase{Problem and approach}}
Applying complex deep learning models to financial time series is limited by data scarcity and non-stationarity of underlying data distributions \cite{mills2008econometric}. Forecasting models, particularly ones based on artificial neural networks like the LSTM used in this study, require large datasets to effectively learn robust temporal dependencies and prevent overfitting \cite{kaplan2020scaling}. Since financial market properties change rapidly, data collected over long periods can become irrelevant. Our approach mitigates this by generating high-quality synthetic data, that faithfully reproduces the characteristics of the limited in size training sets.

\subsection{The generative model}
We selected a TTS-GAN \cite{li2022tts} as the generative model, leveraging the Transformer's self-attention mechanism, which is better than RNNs for modeling the long-range temporal dependencies critical in financial data. 

We acknowledge that the original TTS-GAN architecture has been published elsewhere. However, in order to adopt it for financial domain we contribute by introducing technical enhancements to the implementation and training algorithm necessary for achieving stable convergence:
\begin{enumerate}
    \item \textbf{Optimized Architecture Implementation:} To achieve optimal computational efficiency and training stability, the core Transformer modules (encoder and decoder blocks) were implemented using the highly-optimized native PyTorch \cite{paszke2019pytorch} implementation rather than a custom implementation from original paper\cite{li2022tts}. This substitution significantly improved convergence of the model.

    \item \textbf{Enhanced Training Algorithm:} The adversarial training process was modified to ensure stable learning. Specifically, we improved the training process from the original TTS-GAN's by adding a "Simplified Gradient Penalty," which is calculated only on real samples. This technique was introduced by Mescheder et al. \cite{mescheder2018training}. This turned out to be critical for stabilizing the adversarial balance and accelerating the convergence rate of both the generator and discriminator.
\end{enumerate}

These targeted improvements address the practical challenge of successfully training a generative architecture on limited, volatile data, resulting in the high-fidelity augmented dataset. 

\subsection{The forecasting model}
For the forecasting component, we intentionally employed a standard LSTM network despite the availability of more advanced architectures \cite{kwiatkowski2025comparing}. This choice serves the primary research objective: isolating and quantifying the performance gain achieved solely through the GAN-based data augmentation process.

By focusing on the basic LSTM architecture, we provide a clear validation of the utility of the augmentation technique, independent of specific model-level architectural advantages. In addition, the LSTM architecture is a popular choice for a reference forecasting model, e.g. applied in the research by Fons et al. \cite{fons2020evaluating}, in the research on TimeGAN \cite{yoon2019time}, in the research by Bańka \cite{banka2025applying} and many others. The choice of LSTM for our experiments makes the results of our work more relatable with the results of other research. 

\subsection{Hyper-parameter tuning}
In order to ensure high quality of outcomes, hyper-parameters of both GAN and LSTM models were tuned. 

Hyper-parameter optimization for GAN model required tuning parameters of both generator and discriminator components. We optimized number of transformer encoder blocks (\(D_{g}\) and \(D_{d}\)), embedding size (\(M_{g}\) and \(M_{d}\)), number of attention heads (\(H_{g}\) and \(H_{d}\)) and patch size (\(P_{g}\) and \(P_{d}\)). The optimization process yielded the following parameter values: \(D_g=3\), \(D_d=3\); \(H_g=5\), \(H_d=30\); \(M_g=10\), \(M_d=90\); and \(P_g=15\), \(P_d=15\). A key observation from these results is that the discriminator required significantly higher capacity (specifically in its embedding size, \(M_d=90\), and number of attention heads, \(H_d=30\)) than the generator.

For LSTM model we used  PyTorch's \verb|nn.LSTM| module \cite{paszke2019pytorch}. The optimization process focused on three key hyper-parameters: \textit{hidden\_size}, \textit{num\_layers}, \textit{dropout}. Since the hyper-parameter space was relatively small, we opted for a simple, exhaustive grid search to find  the optimal configuration. This optimization process yielded the following values: a hidden size of 64, three recurrent layer\textbf{s}, and a dropout rate of 0.2.

\subsection{Monitoring the quality of generated data}
An important part of the augmentation process is ensuring that samples used to extend the size of the original dataset are realistic. The samples should exhibit characteristics of the original data samples while not being identical. Achieving this required proper monitoring of the model convergence. In this section, we describe the monitored metrics that allowed us to detect and overcome the inherent challenges associated with GAN training, such as mode collapse, vanishing gradients, and general instability \cite{megahed2024comprehensive}. Detecting these issues was required to ensure the quality and diversity of the synthetic data necessary for effective augmentation. 

External metrics are required to monitor training progress because the GAN loss functions are uninterpretable. The generator loss is calculated on the basis of the number of samples that the discriminator correctly classified. Discriminator loss is calculated on the basis of misclassified samples. The loss of the generator is a function of the discriminator parameters and vice versa\cite{goodfellow2014generative} (equations \ref{eq:2} and \ref{eq:3}). That property makes it impossible to qualify if the model is converging based on values of loss functions.

\begin{equation}\label{eq:2}
    d\_loss = MSE(D(real), 1) + MSE(D(G(z)), 0)
\end{equation} 
\begin{equation}\label{eq:3}
    g\_loss = MSE(D(G(z)), 1)
\end{equation}

To measure the model convergence we used external measures, well established Wasserstein distance \cite{santambrogio2015optimal} and proposed modified version of DeD-iMs \cite{calderon2022dataset} based on the DTW distance. Both of this methods can be used to measure the similarity between the generated synthetic dataset and the real dataset.

\subsubsection{Proposed Dynamic Time Warping based Convergence Metric}
Measuring convergence with a metric based on Dynamic Time Warping (DTW) was necessitated by the inherent limitations of the Wasserstein distance. While Wasserstein distance is applicable to time series, it primarily treats the data as probability distributions, ignoring the crucial chronological order and phase alignment of events that define sequential similarity \cite{cohen2021aligning}. DTW, an elastic dissimilarity measure, is therefore better suited to account for the stretching, permutation, and shifting of important patterns between real and synthetic sequences. However, since we require a measure of similarity between entire datasets, we propose combining DTW with the Deep Dataset Dissimilarity Measure (DeD-iMs \cite{stolte2024methods}). We call this new measure DTW DeD-iMs. We demonstrate this combined approach as a simple yet effective method for comparing time-series datasets, proving particularly useful for monitoring the convergence of the GAN training process.

The DTW \cite{sakoe2003dynamic} \cite{berndt1994using} enables measuring distances between individual samples. The DTW is a well established method that was already tested in many practical applications, like similarities based querying of image databases \cite{bartolini2005warp}, matching shapes for authentication purposes \cite{kholmatov2005identity} and many others. DTW is an elastic dissimilarity measure which works by minimizing the cost of alignment between two time series. The measure can be described with the following formulas:

\begin{equation}
d_{DTW}(x, y) = D_{M,N}(x,y)
\end{equation}

Where \(x\) and \(y\) are time series of \(M\) and \(N\) points accordingly and \(D_{M,N}(x,y)\) is described as:

\begin{equation}
D_{i,j}(x, y) = f(x_{i}, y_{j}) + min \begin{cases} 
& D_{i,j-1}(x,y)
\\ 
& D_{i-1,j}(x,y)
\\
& D_{i-1,j-1}(x,y)
\end{cases}
\end{equation}

We used \(f(x_{i}, y_{j}) = (x_{i} - y_{j})^{2}\), which is the most common choice for a cost function.

The Deep Dataset Dissimilarity Measure \cite{stolte2024methods}\cite{calderon2022dataset} is the second component required to measure similarity between datasets. By providing a framework for comparing entire sets, DeD-iMs complements the base distance metric—in this case, DTW—which measures similarity only between individual samples. The process for measuring similarity between two sets, \(S_{a}\) and \(S_{b}\), involves the following steps:

\begin{enumerate}
\item Draw a random sample of size \(n\) from sets \(S_{a}\) and \(S_{b}\): \(S_{a}^{k}\) and \(S_{b}^{k}\).

\item Calculate \(d_{i} = min_{k} \|s_{i} - s_{k}\|_{d}\) for each sample \(s_{i} \in S_{a}\), where \(s_{k} \in S_{b}\) is the closest element to \(s_{i} \in S_{a}\) measured with measure \(d\). In our implementation the distance between each pair from \(S_{a}\) and \(S_{b}\) was calculated using the DTW. This yields a list of \(n\) distances \(D_{d}\).

\item Calculate a reference list of distances for the same samples in the dataset \(S_{a}\) from itself - \(D'_{d}\).

\item The value of the dissimilarity measurement is an average of absolute differences between reference and inter-dataset distances.
\end{enumerate}

\begin{figure}
    \centering
    \includegraphics[width=1\linewidth]{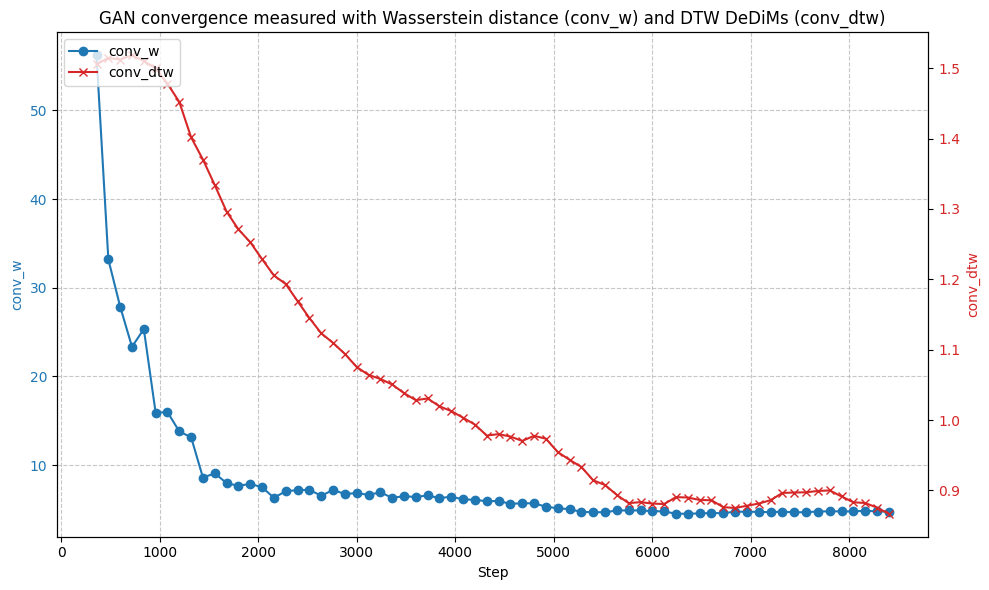}
    \caption{Comparison between TTS-GAN convergence measured using Wasserstein distance and DTW DeDiMs metric}
    \label{fig:dtw-dedims}
\end{figure}

Figure \ref{fig:dtw-dedims} plots the values of the convergence metric, clearly showing that both metrics converge during training. However, when comparing this result to the convergence measured using the Wasserstein distance, a key difference emerges: the DTW DeD-iMs metric continues to decrease through the later epochs. This behavior suggests that relying on DTW enabled the detection of subtle improvements in the temporal fidelity of the generated sequences, even after the overall feature distribution had converged.

Using Wasserstein distance and our proposed DTW DeD-iMs ensured high quality of data used for augmentation which was necessary to validate our core hypothesis.

\section{\uppercase{Experimental Design and Procedure}}
The primary objective of this study is to quantify the impact of utilizing GAN-augmented data on the forecasting performance of a deep learning model. We hypothesize that training on an augmented dataset will yield a lower generalization error compared to training exclusively on real data. To verify this, we propose a three-phase procedure:

\begin{enumerate}
\item \textbf{Generative Model Training}: A TTS-GAN is trained on real segments of the stock price time series to generate high-fidelity synthetic data.
\item \textbf{Baseline Forecasting}: The LSTM forecasting model is trained and evaluated using only real samples from the financial datasets.
\item \textbf{Augmented Forecasting}: The same LSTM model is trained and evaluated using an augmented dataset, which combines the original real samples with the synthetic samples generated by the trained GAN.
\end{enumerate}

To account for the non-stationarity and regime shifts inherent in financial time series, the entire experiment was conducted on 40 distinct datasets, each comprising price data from different, non-overlapping time periods. This approach allows us to test the augmentation method's generalization ability across diverse market conditions. We used historical daily closing prices for Bitcoin (last 10 years) and the S\&P500 Index (1990--2025), sourced from publicly available sources.

\begin{enumerate}
    \item \textbf{Preprocessing}: Data were initially smoothed using a backward-looking exponential moving average to mitigate the influence of transient outliers and high-frequency noise.
    \item \textbf{Sample Generation}: The time series were divided into overlapping sequential samples, each consisting of $K$ time points. We prepared datasets with sequence lengths $K = 90$ and $K = 120$. The forecasting model predicts $S$ data points based on $T$ observations, where $K = T + S$. For $K = 90$, $T$ was set to $60$ and $S$ to $30$ and for $K = 120$, $T = 80$ and $S = 40$. The resulting sample counts across the 40 time windows were:
    \begin{itemize}
        \item 10x Bitcoin, $K = 90$: $\approx 370$ samples each
        \item 10x Bitcoin, $K = 120$: $\approx 340$ samples each
        \item 10x S\&P500, $K = 90$: $\approx 890$ samples each
        \item 10x S\&P500, $K = 120$: $\approx 860$ samples each
    \end{itemize}
    \item \textbf{Sample-wise Normalization}: Each $K$-point sample was normalized independently using the MinMaxScaler transformation:
    \begin{equation}
    X_{\text{scaled}} = \frac{(X - X_{\min})}{(X_{\max} - X_{\min})}
    \end{equation}
To prevent information leakage from the future, the $X_{\max}$ and $X_{\min}$ values were calculated solely based on the initial observation portion ($T$ data points) of the sample, not the future forecast window ($S$).
\end{enumerate}

Each dataset was split into a training set ($T$), validation set ($V$), and test set ($E$) using the proportions $T$: $60\%$, $V$: $20\%$, and $E$: $20\%$. The GAN was trained on $T \cup V$ and never shown any samples from the test set $E$. For the augmented experiments, we employed a $1:1$ augmentation ratio, doubling the size of the training set ($|T|_{\text{augmented}} = 2|T|$).

\section{\uppercase{Results and discussion}}
To validate the hypothesis stated in this study, we measured the improvement in forecasting error across 40 different datasets. We trained an LSTM on both original and augmented datasets, resulting in 40 pairs of forecasting models. For each pair, we calculated the MSE on the test set and computed the difference between the model trained on original data and the one trained on augmented data. Next, we grouped these results by dataset type (e.g., all Bitcoin datasets consisting of samples of length 90). The mean MSE improvement was calculated, as well as the standard error (SE) and a paired t-test (p-value). Aggregated results are presented in Table \ref{tab:sp500_augmented_vs_real}.

\begin{table*}[t]
    \centering
    \begin{tabular}{|c|c|c|c|c|} \hline 
         dataset & sequence length& Mean MSE Improvement& SE & p-value \\ \hline 
         Bitcoin & 90 & 0.102 & 0.024 & 0.002\\ \hline 
         Bitcoin & 120 & 0.132 & 0.016 & 0.003\\ \hline 
         SP500 & 90 & 0.037 & 0.01 & 0.006 \\ \hline 
         SP500 & 120 & 0.044 & 0.018 & 0.007\\ \hline
    \end{tabular}
\caption{Forecasting MSE Improvements: Augmented vs. Real Dataset}
\label{tab:sp500_augmented_vs_real}
\end{table*}

Across all four experimental conditions, augmentation led to a reduction in the average MSE compared to the baseline. Low Standard Error (SE) indicates that the sample mean is a highly reliable estimate of the true population mean. The consistently low SE values in our experiments suggest that measured results are reliable and reproducible. Specifically, a p-value below 0.05 indicates statistical significance of these results and confirms stated hypothesis.

Larger improvements in MSE occurred for the Bitcoin dataset at both K=120 and K=90. Given that the Bitcoin dataset was smaller than the S\&P500 data, this result highlights that augmentation provides more performance gains when original dataset is limited.

While the MSE improvements may appear modest in absolute terms in some conditions, their consistency and reproducibility across 40 distinct time windows are highly significant. In the high-stakes domain of financial forecasting, even marginal, consistent improvements in predictive accuracy can translate to substantial practical advantages. Furthermore, the positive impact was more pronounced on the Bitcoin dataset, which is characterized by higher volatility and smaller effective dataset size compared to the S\&P500. This outcome aligns with our theoretical premise: the more data-scarce and volatile the domain, the greater the benefit derived from high-fidelity data augmentation.

The experimental design, which utilized 40 time windows for training GANs and LSTMs, was critical for implicitly testing the model's resilience across various market regimes (e.g., bull vs. bear markets, periods of high vs. low volatility).

While working on verifying our hypothesis we contribute with technical enhancements to the TTS-GAN implementation and training strategy. This leads to demonstrably faster and more stable convergence compared to the original architecture. On top of that, we address a critical challenge in GAN training, the lack of consensus on metrics for evaluating generated data specific to time series \cite{brophy2023generative} \cite{jeng2025generative}. We introduce a novel dataset similarity measure adapted for time series that combines Dynamic Time Warping (DTW) \cite{berndt1994using}, to assess temporal fidelity, and the Deep Dataset Dissimilarity Measure (DeD-iMs) \cite{calderon2022dataset}, to assess overall distribution similarity.

The introduction of the combined DTW DeD-iMs metric provided a more signal for monitoring the GAN training progress. We observed that during training (referenced in Fig. \ref{fig:dtw-dedims}), the Wasserstein distance converged and flattened prematurely, suggesting that the model had achieved distributional similarity but providing no further guidance on improving temporal quality. In contrast, the DTW DeD-iMs metric continued to decrease linearly over later epochs.

The promising results from this empirical study warrant several directions for future exploration. While we focused on the LSTM to isolate the effect of augmentation, subsequent research should include more complex deep learning architectures, such as transformer-based architectures, which have recently received more attention in financial forecasting field \cite{szydlowski2024hidformer}. Given that these larger models are more data-demanding and prone to overfitting, the augmentation technique may potentially yield even greater performance gains.

Finally, we believe that our research opens venues for more practical applications. One such application could be an adaptive system for market regime shift detection. A main focus would be to investigate the potential of the discriminator component to detect shifts in the underlying domain, such as market regime changes. Successful application could enable the implementation of a fully automated, self-adapting machine learning solution. In such a system, the discriminator would detect a regime change and trigger the re-training of the generator for data augmentation and, subsequently, the training of the forecasting model on augmented dataset. This application of the GAN leveraging unsupervised learning could lead to a system that operates with limited need for human intervention.

\section{\uppercase{Conclusion}}
The research successfully validated the use of GAN-based data augmentation for enhancing deep learning forecasting capabilities in financial time series. We utilized an enhanced Transformer-based GAN (TTS-GAN), which included key implementation and algorithmic modifications necessary for achieving stable convergence on volatile, non-stationary financial data.

The core finding is that training a standard LSTM forecasting model on augmented data, generated by our enhanced TTS-GAN, reduces MSE across 40 distinct datasets and various forecasting horizons. This validates the utility of our high-fidelity generative approach for mitigating the critical challenges of data scarcity and regime shifts in financial markets.

Furthermore, we contributed by introducing a novel convergence monitoring metric that combines DTW and the DeD-iMs. This metric was shown to be highly effective at assessing both the distributional and temporal fidelity of synthetic sequences, providing crucial insight into the training process that standard methods like Wasserstein distance often fail to capture.

\bibliographystyle{apalike}
{\small
\bibliography{bibliography}}

\end{document}